# Integration of Object Detection and Small VLMs for Construction Safety Hazard Identification

Muhammad Adil, Mehmood Ahmed, Muhammad Aqib, Vicente A. Gonzalez, Gaang Lee, Qipei Mei*

Infrastructure Human Tech (IHT) Lab, Department of Civil and Environmental Engineering, University of Alberta, Edmonton, Alberta, Canada

*Corresponding author's email address: qipei.mei@ualberta.ca

**Abstract**

Accurate and timely identification of construction hazards around workers is essential for preventing workplace accidents. While large vision–language models (VLMs) demonstrate strong contextual reasoning capabilities, their high computational requirements limit their applicability in near real-time construction hazard detection. In contrast, small vision–language models (sVLMs) with fewer than 4 billion parameters offer improved efficiency but often suffer from reduced accuracy and hallucination when analyzing complex construction scenes. To address this trade-off, this study proposes a detection-guided sVLM framework that integrates object detection with multimodal reasoning for contextual hazard identification. The framework first employs a YOLOv11n detector to localize workers and construction machinery within the scene. The detected entities are then embedded into structured prompts to guide the reasoning process of sVLMs, enabling spatially grounded hazard assessment. Within this framework, six sVLMs (Gemma-3 4B, Qwen-3-VL 2B/4B, InternVL-3 1B/2B, and SmolVLM-2B) were evaluated in zero-shot settings on a curated dataset of construction site images with hazard annotations and explanatory rationales. The proposed approach consistently improved hazard detection performance across all models. The best-performing model, Gemma-3 4B, achieved an F1-score of 50.6%, compared to 34.5% in the baseline configuration. Explanation quality also improved significantly, with BERTScore F1 increasing from 0.61 to 0.82. Despite incorporating object detection, the framework introduces minimal overhead, adding only 2.5 ms per image during inference. These results demonstrate that integrating lightweight object detection with small VLM reasoning provides an effective and efficient solution for context-aware construction safety hazard detection.

**Keywords:** Object detection, Vision–language models, Hazard detection, Real-time monitoring

## 1. Introduction

The construction industry is ranked as one of the most hazardous industries globally. In Canada, although it represents only about 7.8% of the national workforce, it accounted for 17.3% of all occupational fatalities in 2023 [1]. Similarly, in the U.S., statistics indicate that approximately 20% of work-related deaths occurred within the construction sector [2]. The three leading causes of fatal accidents in construction are slips, trips and falls, transportation-related incidents, and improper contact with objects or equipment [2]. The high frequency and severity of these events, highlight the need to address challenges in timely identification of hazards on construction sites.

Construction sites are characterized by complexity, with hazards that are spatiotemporally dynamic due to the continuous movement and interaction of workers and machinery, as well as the evolving nature of tasks throughout the course of a project [3]. Conventional safety monitoring approaches fall short in effectively identifying hazards, as they are limited by their reliance on manual observation, which is time-consuming, prone to bias, and often results in inefficiency and errors [4,5]. To address the limitations associated with manual inspection and automate the observation of construction sites, computer vision (CV) technologies have been adopted [6]. Object detection is a CV method that can extract and localize predefined categories of objects within construction images [7]. Previous studies have explored applications of object detection to detect safety hazards from construction images by detecting hazard-related objects such as workers, personal protective equipment (PPE), machinery, and dangerous materials [8–11]. However, effective hazard detection requires not only recognizing objects but also interpreting their context and semantic relationships [12]. For example, detecting machinery alone does not necessarily indicate danger; the hazard arises when its proximity to workers suggests a potential collision risk. The 'trace intersecting theory' reinforces this perspective, arguing that accidents typically arise when unsafe worker actions and unsafe site conditions intersect in the same time and space [13,14]. Therefore, hazard detection systems must go beyond object-level identification to reason about the broader scene. By analyzing semantic relationships among elements, unstructured visual data can be transformed into structured textual representations, which can then be combined with safety regulations to enable more advanced and reliable hazard identification [14]. This gap has motivated researchers to explore more advanced AI systems capable of understanding semantics and context.

Vision–language models (VLMs) combine visual and textual information by employing a pre-trained visual encoder to extract image features, a cross-modal alignment module to fuse them with text representations, and an autoregressive decoder to generate outputs [15]. This architecture enables VLMs to move beyond simple object-level recognition, allowing them to interpret complex scene structures, infer semantic relationships, and produce context-aware representations of unstructured visual data. Recent studies have demonstrated the potential of VLMs for construction activity monitoring and hazard detection [16–21]. Despite the promise, existing VLM-based approaches face limitations. First, large VLMs used in these studies such as GPT-4o, Llama 3.2 11B, Qwen 2-VL, InternVL2 8B deliver high accuracy but demand substantial computational resources, making them impractical for deployment on construction sites where near real-time performance is critical and hardware resources may be limited [22]. Second, small vision language models (sVLM), although computationally efficient, suffer from less accuracy, generating false or irrelevant hazard predictions [20]. This trade-off between efficiency and accuracy remains a major barrier to the adoption of VLMs in practical construction safety hazard detection.

To address these challenges, this study introduces a novel integration framework that combines object detection with sVLM for efficient hazards identification around workers. The framework leverages an object detection model to localize workers and machinery, and the resulting detections are encoded into structured prompts that guide the sVLM to analyze hazards in the vicinity of workers. The contributions of this study are threefold. First, it proposes a detection-guided framework that integrates object detection with small VLMs for worker-centric hazard identification. Second, it provides a comparative evaluation of multiple sVLMs in zero-shot settings for construction safety tasks. Third, it demonstrates that the proposed approach improves hazard detection performance and reasoning of the sVLMs with minimal computational overhead.

Section 2 reviews the related work. The proposed methodology is presented in Section 3, followed by the experimental setup and evaluation metrics in Section 4. Results are reported in Section 5, and Section 6 discusses the key findings. Finally, the conclusion of this proposed study is presented in Section 7.

## 2. Related Works

### 2.1. Object Detection for Construction Safety Monitoring

Object detection is a computer vision technique that identifies and locates the entities in an image. Early implementations of object detection for construction monitoring used techniques such as subtraction algorithms [23] and machine learning models [24]. With development of more sophisticated deep learning methods, recent studies have explored methods based on convolutional neural networks (CNN) and recurrent neural networks (RNN) for automated safety monitoring [6]. Algorithms such as Single Shot MultiBox Detector (SSD) [25], You

Only Look Once (YOLO) [26] , and Faster R CNN [27] have been adopted to identify workers, equipment, and PPE in construction environments. For example, Wu et al. [28] proposed a one stage CNN based on SSD to monitor hardhat compliance. Kim et al. [29] proposed a drone-assisted monitoring method that uses third version of YOLO to detect struck-by hazards. Son et al. [30] adopted fourth version of YOLO to track workers on construction site for . Lee at al. used YOLOv5 to identify tools in indoor construction sites. Zhang et al. [31] developed a safety management system by combining computer vision based on Fast R-CNN and a location system to track and alert workers in case of a hazardous situation. Beyond worker tracking and PPE, object detection has also been applied to identify other hazards such as falls [32,33] and fire risks [34] on-site. While object detection has achieved significant progress in construction safety applications, its capabilities remain limited to identifying predefined objects and their locations. It lacks the ability to reason about contextual relationships between entities, which is essential for assessing complex hazards. These limitations restrict its effectiveness in performing high-level semantic understanding of construction scenes.

*2.2. Vision Language Models for Construction Safety Hazard Detection*

To overcome the above limitations, recent works leverage VLMs that combine visual and textual understanding. By combining visual encoders with the contextual reasoning capabilities of Large Language Models (LLMs), VLMs enable processing of images and text for multimodal understanding [35]. This integration allows models to move beyond object-level recognition, interpreting complex scene structures, inferring relationships, and generating context-aware representations of unstructured visual data. Early breakthroughs in this domain include the CLIP model developed by Radford et al. [36] which leveraged contrastive learning to align images with corresponding text descriptions, laying the foundation for cross-modal understanding. Building on this progress, subsequent models, including Flamingo [37], BLIP [38], and LLaVA [39] have extended beyond simple image-text matching to support generative tasks. This shift transformed VLMs from simple classifiers to interactive visual assistants capable of following user prompts and generating contextual responses. More recent large VLMs such as GPT-4 [40], Gemini [41], LLaMA [42], Qwen [43], and InternVL [44] have demonstrated remarkable performance in tasks such as image captioning, visual question answering, and object grounding.

VLM's ability to comprehend both visual and textual inputs, combined with their prompt-following capabilities, makes them promising tools for conducting construction inspections [20]. The key idea is to inject semantic knowledge (e.g. safety rules, hazard descriptions) into the vision system so that hazards can be recognized not just by object classes, but by their context and meaning. For example, Tsai et al. [18] developed an automated safety inspection system by fine-tuning CLIP [45] with a prefix captioning mechanism tailored to construction safety contexts. Ding et al. [46] proposed a visual question answering (VQA) system for identifying unsafe behaviors using a Vision-and-Language Transformer (ViLT) model. Fan et al. [47] proposed ErgoChat, that uses ViT combined with LlaMA2 to assess the postural ergonomic risk of construction workers. With the development of large pre-trained models, zero- and few-shot learning has become feasible for construction hazard detection. Chen et al. proposed Clip2Safety that uses frozen CLIP model for PPE detection [17]. Adil et al. [20] proposed a large pre-trained VLM based framework that employs prompt engineering to embed safety guidelines into queries, enabling the model to detect both general and, context-specific hazards on construction sites. Chen and Zou [48] conducted an evaluation of current state-of-the-art large pretrained VLMs in zero-shot and few-shot settings. Chen and Yin [19] introduced ChatCH, that uses Qwen2-VL to identify potential hazards and generate assessment reports.

Despite these advancements, the deployment of VLMs in construction safety hazard detection faces two key challenges. First, the use of commercial VLMs, such as GPT-4 or Gemini, is constrained by cost, privacy, and security concerns due to their closed-source nature [49]. These models rely on web-based APIs, which makes handling sensitive site information difficult in the absence of clear regulatory frameworks [49]. Second, the high computational demands of large VLMs hinder near real-time, on-site deployment [50]. While open-source VLMs support on-premises operation, they typically require extensive GPU resources, for example, multiple A100 or H100 GPUs to process multi-camera inputs, far exceeding the capacity of standard desktop systems. These challenges highlight the need for lighter, resource-efficient alternatives.

A practical solution can be the use of open-sourced sVLM with fewer than four billion parameters [51]. These lightweight models can be deployed on a wider range of hardware, from embedded edge devices such as the Jetson AGX Orin to commodity GPUs like the RTX 5070, making them more practical for field deployment. Some notable sVLMs include Gemma-3 4B [52], Qwen-3-VL 2B/4B [43], InternVL-3 1B/2B [44], and SmolVLM 2B [53], which balance efficiency with reasonable performance on multimodal benchmarks. These models have shown potential for on-site deployment by reducing inference latency and lowering hardware requirements, making them attractive for field applications. However, their reduced scale comes with a trade-off. Small VLMs typically achieve lower accuracy [54], particularly in complex or cluttered construction scenes, and tend to lose focus when multiple hazards or contextual cues must be considered simultaneously. Furthermore, their limited capacity makes them more prone to hallucinations or omission errors [54], where critical hazards are either misidentified or overlooked. Addressing this trade-off, between efficiency and reliability, remains a critical research gap for enabling scalable, real-time hazard detection on construction sites.

*2.3. Knowledge Gaps and Research Objectives*

Despite advancements in digital monitoring and computer vision for construction safety hazard detection, the following knowledge gaps persist:

(1) VLMs have shown significant potential in contextual hazard identification. Recent studies demonstrate the potential of VLMs for PPE compliance[17], hazards identification [20], and automated report generation [19]. However, research has so far been limited to proof-of-concept studies, and there has been little systematic evaluation of their feasibility for on-site, near real-time deployment. The significant computational demands of large VLMs make them impractical in field conditions, yet there is limited literature explicitly addressing this deployment barrier [20].

(2) sVLMs, with fewer than four billion parameters, are efficient alternatives, deployable on edge devices [54]. These models have less inference latency and hardware requirements, enabling potential field deployments. Despite their potential, their performance tradeoffs, such as lower accuracy, reduced focus in cluttered scenes, and higher susceptibility to omission errors [54] remains underexplored in the construction safety domain. In particular, there is limited research on methods to enhance the accuracy and robustness of sVLMs for hazard detection tasks.

Based on these research gaps, this study aims to enhance the accuracy and reliability of sVLMs for construction safety hazard detection by an integration framework that combines object detection with sVLMs. The validation focuses on (i) evaluating representative sVLMs, including Gemma-3 2B, Qwen-3 2B, and InternVL-3 1B/2B, in zero-shot settings on a curated dataset of construction images to establish baseline performance; (ii) systematically comparing the performance of these baseline sVLMs with the proposed detection-guided framework to assess improvements in hazard identification and rationale generation; and (iii) addressing the trade-off between computational efficiency and contextual accuracy, thereby advancing the development of lightweight yet effective safety hazard detection systems capable of supporting proactive hazard detection in dynamic construction environments.

**3. Methodology**

This study presents an integrated object detection and VLM framework for near real-time safety hazard detection on construction sites. The framework is designed to combine the spatial precision of object detection with the contextual reasoning ability of sVLM. The overview of the proposed framework is illustrated in Figure 1.

The proposed framework comprises two primary components: (1) an object detection module responsible for identifying and localising workers and machinery, and (2) a detection-guided sVLM module that performs contextual hazard assessment based on structured detection outputs and predefined safety criteria.

Given an input construction site image, the object detection module first extracts spatially grounded information by producing two primary outputs: (1) bounding box coordinates specifying the location of each detected entity, and (2) class labels that categorize each detection into predefined classes. These structured outputs explicitly identify which objects are present in the scene and where they are located. The detection results are subsequently transformed into a structured natural language prompt. This prompt embeds entity identities, approximate spatial information, and predefined hazard assessment rules. The prompt, together with the original image, is then provided to the sVLM. The sVLM performs multimodal reasoning and generates structured hazard assessments, including hazard categories and explanatory justifications. This design preserves global scene understanding while anchoring hazard detection to spatially grounded entities, thereby reducing ambiguity and improving the accuracy of sVLM.

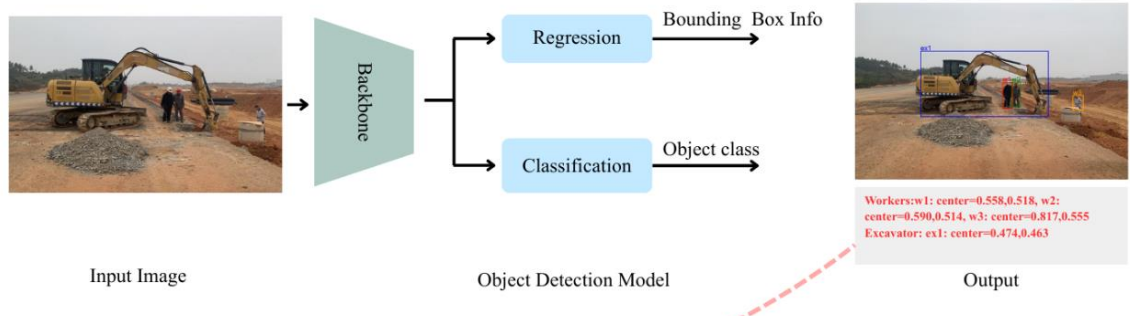

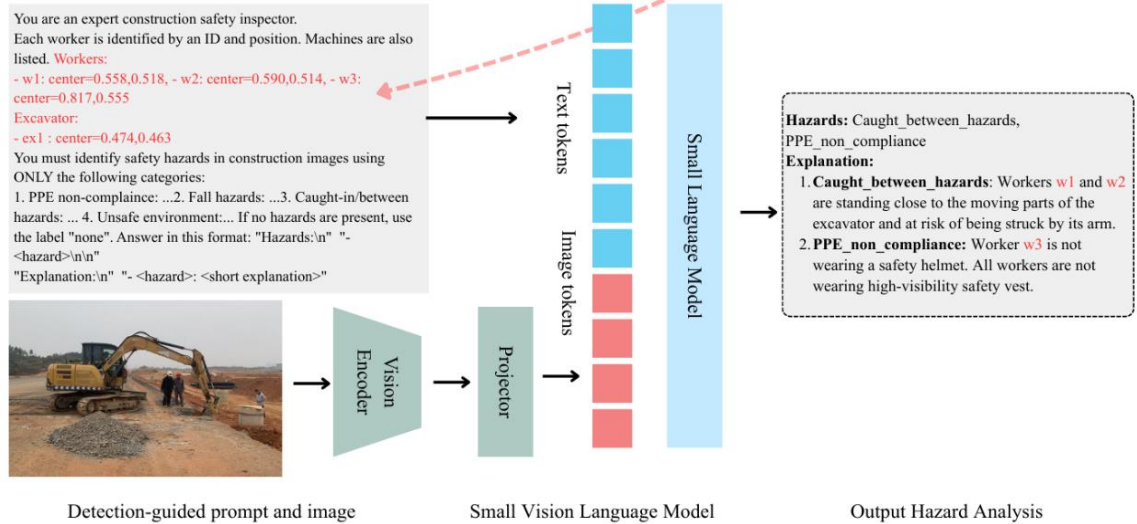

Figure 1 – Overview of proposed framework

This study follows a systematic methodology to evaluate the proposed framework. The overview of the methodology is illustrated in Figure 2. In the first step of study, we compile an object detection dataset, divide the dataset into training, validation and test sets. We then train the object detection model. In the second stage, we collect construction images and label the images for hazards. Subsequently, a framework integrating the training object detection model and a pre-trained sVLM is developed, and multiple sVLMs are evaluated using the developed dataset. Finally, to demonstrate the practical application of the framework, the developed framework is used to automatically monitor the site in near real-time. Detailed description of the stages is provided in the following sections.

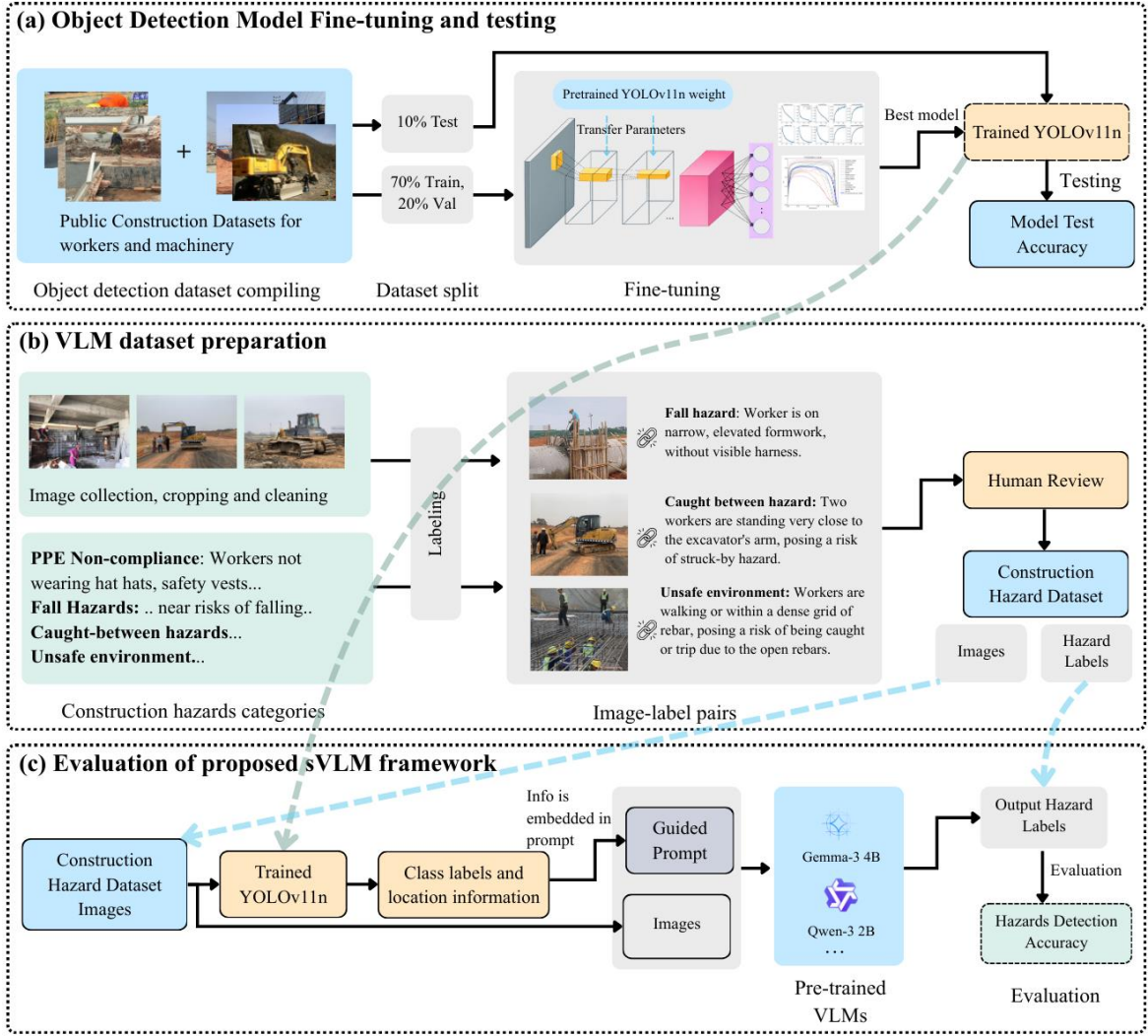

Figure 2 – The overall methodology of this study

*3.1. Object Detection*

The object detection module serves as the spatial grounding component of the framework. Its primary function is to localise and classify workers and machinery within construction site images.

Given an input image $I \in \mathbb{R}^{H \times W \times 3}$, where *H and W* represent the image height and width, and the three channels correspond to the RGB colour space. The object detection model processes the input image and produces a set of detections defined as

$$D = \{(b_i, c_i, s_i)\}_{i=1}^{N} \tag{1}$$

where $N$ denotes the number of detected objects in the image, $b_i$ represents the bounding box parameters of the $i^{th}$ *detection*, $c_i$ denotes the predicted class label, and $s_i$ denotes the confidence score associated with the prediction. The bounding box $b_i$ is defined as

$$b_i = (x_i, y_i, w_i, h_i) \tag{2}$$

where $(x_i, y_i)$ represent the centre coordinates of the bounding box relative to the image dimensions, and $w_i$ and $h_i$ represent the width and height of the bounding box, respectively. These parameters are expressed in normalized image coordinates, allowing the representation to remain independent of the original image resolution. The predicted class label $c_i$ belongs to a predefined set of object classes. To facilitate structured reasoning in the subsequent module, each detected entity is assigned a unique identifier. Workers are labelled as $W = \{w_1, w_2, \ldots, w_m\}$ and machinery instances are labelled according to their class type with sequential indices (e.g., excavators as $ex_1, ex_2$, dump trucks as $dt_1$, etc.). The indexing is determined based on the horizontal spatial ordering of detections within the image. Formally, detections are sorted according to their horizontal centre coordinate

$$x_{(1)} \leq x_{(2)} \leq \cdots \leq x_{(N)} \tag{3}$$

where $x_{(i)}$ represents the ordered centre coordinate of the $i^{th}$ detection. Identifiers are then assigned sequentially from left to right. This ordering ensures a deterministic mapping between detected entities and their textual representations in the subsequent reasoning stage.

3.1.1. Dataset compilation and preprocessing

The object detection task was carried out using the YOLOv11n algorithm due to its balance between computational efficiency and detection accuracy. The model was trained using a dataset sourced from publicly available safety datasets, including the Construction Site Safety Dataset [55] for workers and the Alberta Construction Image Dataset (ACID) [56] for machinery. The combined dataset comprised a total of 10,568 annotated images, with classes including workers and ten categories of heavy machinery, as detailed in Table 1.

Table 1. Object detection classes

| Class | No. of Annotations |
|---|---|
| Worker | 3527 |
| Cement Truck | 655 |
| Compactor | 877 |
| Dozer | 775 |
| Dump Truck | 948 |
| Excavator | 922 |
| Grader | 940 |
| Mobile Crane | 898 |
| Tower Crane | 363 |
| Wheel Loader | 949 |
| Backhoe Loader | 978 |

The dataset was divided into three subsets: 70% for training, 20% for validation, and 10% for testing. All images from both datasets were preprocessed prior to training and evaluation by resizing them to a standardized resolution of 640 × 640 × 3 pixels in the RGB color space, ensuring consistency and enabling reliable performance comparison.

3.1.2. Training and validation configuration

The experimental settings define the computational environment, hyperparameter configuration, and optimisation strategies employed during model training. Table 2 presents the hyperparameters used for the object detection model. The YOLOv11n model was implemented using the Ultralytics framework (version 8.3.127) with Python 3.11 and PyTorch 2.7.0, accelerated by CUDA 12.6. Training was conducted on an NVIDIA RTX 4080 GPU (16 GB memory). The model was trained for 50 epochs with a batch size of 16 and an input resolution of 640 × 640 pixels. Deterministic training was enabled to ensure consistent results across runs. The optimisation process used Stochastic Gradient Descent (SGD) with an initial learning rate of 0.01, a final learning rate fraction (lrf) of 0.01, momentum of 0.937, and weight decay of 0.0005. A warm-up phase of 3 epochs was applied, with a warm-up momentum of 0.8 and a warm-up bias learning rate of 0.1 to stabilise early-stage training. . To mitigate overfitting and improve generalisation, multiple augmentation techniques were applied. Mosaic augmentation was enabled (probability = 1.0) to improve multi-scale object detection. Horizontal flipping was applied with a probability of 50 %, translation up to ±10 %, and scaling up to 0.5 were used to simulate spatial variability. Colour augmentations were introduced through HSV adjustments, with hue variation of 0.015, saturation of 0.7, and brightness of 0.4. Random erasing was applied with a probability of 0.4 to simulate occlusions, and RandAugment-based automatic augmentation further diversified the training data. Validation was performed during training, and an early stopping strategy with a patience value of 15 epochs was configured to prevent excessive training if performance plateaued. The best-performing model weights were automatically saved based on validation metrics.

During the model training process, performance was evaluated using the validation dataset to monitor convergence and assess generalisation capability. Validation metrics were computed after each epoch to track improvements in detection accuracy and localisation performance. This continuous monitoring enabled early identification of overfitting and ensured that the best-performing model weights were retained based on validation results.

Table 2. Experimental Configuration and Key Hyperparameters

| Category | Parameter | Value |
|---|---|---|
| **Environment** | Framework / Library | Ultralytics 8.3.127, PyTorch 2.7.0 (CUDA 12.6) |
|  | GPU | NVIDIA RTX 4080 (16 GB) |
| **Model Setup** | Architecture | YOLOv11n (pretrained) |
|  | Number of classes | 16 |
|  | Image size | 640 × 640 |
| **Training** | Epochs / Batch size | 50 / 16 |
|  | Optimizer | SGD |
|  | Learning rate (initial) | 0.01 |
|  | Momentum | 0.937 |
|  | Weight decay | 0.0005 |
| **Loss Configuration** | Box / Cls / DFL gains | 7.5 / 0.5 / 1.5 |
|  | IoU threshold | 0.7 |
| **Augmentation** | Mosaic / Flip | 1.0 / 0.5 |
|  | HSV (H,S,V) | 0.015 / 0.7 / 0.4 |
|  | Random erasing | 0.4 |
| **Regularisation** | Early stopping patience | 15 |

*3.2 Detection-guided sVLM for hazard detection*

While the object detection module provides spatial grounding through explicit localisation and classification of entities, it does not capture higher-level contextual relationships required for full safety assessment. To address this limitation, a sVLM is used for structured contextual hazard identification. By embedding both spatial entity information and hazard definitions within the prompt, the model receives explicit contextual cues regarding which relationships should be evaluated. Let the filtered detection set produced by the object detection module be denoted as

$$D' = \{(b_i, c_i, s_i, id_i)\}_{i=1}^{M} \tag{4}$$

where $b_i = (x_i, y_i, w_i, h_i)$ represents the bounding box parameters, $c_i$ denotes the predicted class label, $s_i$ denotes the detection confidence score, and $id_i$ represents the assigned identifier for the detected entity. As described in Section 3.2, workers are labelled as $w_1, w_2, ...,$ while machinery instances are labelled according to their type and spatial ordering (e.g., $ex_1$ for excavator, $dt_1$ for dump truck). These identifiers provide a deterministic reference system for the reasoning process.

3.2.1. Detection-to-Prompt Encoding Strategy

The first step in the contextual hazard detection converts the structured detection outputs into a natural language prompt that can be interpreted by the sVLM. Each detected entity is represented in textual form using its identifier, class label, and spatial coordinates derived from the bounding box centre $(x_i, y_i)$. This process transforms the geometric detection representation into a structured textual description. Formally, the detection set $D'$ is transformed into a textual representation

$$T(D') = \{(id_i, c_i, x_i, y_i)\}_{i=1}^{M} \tag{5}$$

which lists all entities present in the scene together with their spatial positions. The textual representation is then embedded within a predefined prompt template that includes both entity descriptions and safety evaluation instructions. The prompt also incorporates predefined safety assessment categories, as presented in Table 3, considered in this study:

Table 3 - Contextual hazard categories used in this study

| Hazard Category | Explanation |
|---|---|
| **PPE Non-Compliance** | Look for Workers not wearing appropriate personal protective equipment. |
| **Fall Hazard** | Look for workers near risks of falling, including at elevated work areas, near open excavations, on temporary structures (ladders, lifts, scaffolding) without safety harness. |
| **Caught-between Hazard** | Risks of workers being struck by, caught, crushed, or pinned by or between machinery, moving objects, structures, confined spaces, or trenches. |
| **Unsafe Environment** | Look for unsafe site conditions such as exposed rebar, uneven terrain, debris or waste, open electrical wires, standing water, poor lighting etc. |

3.2.2. Contextual Hazard Identification

After prompt construction, the full input image $I$ and the detection-guided prompt $P(D')$ are jointly provided to the sVLM. The model processes the visual input through its vision encoder to extract image feature representations, while the textual prompt is processed by the language encoder to generate contextual embeddings.

The inclusion of detection-derived entity information within the prompt anchors the hazard detection process to the spatial structure of the scene. Rather than performing unconstrained scene interpretation, the model focuses on relationships between detected workers, machinery, and environmental elements. This conditioning improves the ability of the model to evaluate safety conditions such as unsafe worker–machinery proximity, missing personal protective equipment, or hazardous environmental configurations.

The proposed framework is model-agnostic and can be integrated with different sVLMs capable of joint visual–textual reasoning. In this study, several open-source sVLM architectures were evaluated within the framework, including Gemma-3 (4B), Qwen-3 (2B and 4B), and InternVL-3 (1B, 2B, and 4B) models. These models were selected due to their relatively small parameter sizes and demonstrated capability for multimodal reasoning tasks. The use of multiple sVLM architectures allows evaluation of the proposed detection-guided prompting strategy across different model families and parameter scales. Detailed implementation settings and experimental comparisons are presented in Section 4.

## 4. Evaluation

To validate the proposed detection-guided framework, a comprehensive experimental evaluation was conducted. The evaluation framework was designed to assess three key aspects of the system: (1) assessing the performance of the object detection module, (2) evaluating the contextual hazard detection capability of the vision–language models, and (3) measuring the quality of the generated reasoning explanations.

*4.1. Evaluation of object detection*

The object detection module was evaluated using the held-out 10% test split of the dataset described in Section 3.2, which contains annotated workers and machinery instances. The evaluation aims to assess the ability of the detector to accurately localize and classify relevant construction entities.

4.1.1. Evaluation Metrics

The primary objective of this evaluation is to measure how effectively the model detects entities in construction scenes. For this task, mean Average Precision (mAP) is a widely used metric that evaluates the accuracy of predicted bounding boxes across various classes and intersection-over-union (IoU) thresholds.

IoU is defined as the ratio between the area of overlap and the area of union of the predicted (PD) and ground truth (GT) bounding boxes:

$$IoU = \frac{area\ (GT \cap PD)}{area\ (GT \cup PD)} \tag{6}$$

An IoU threshold $\alpha \in [0,1]$ is used to determine detection correctness. A predicted bounding box is considered a true positive (TP) if the IoU exceeds the threshold and the predicted class label matches the ground truth. Predictions below the threshold are considered false positives (FP), while ground-truth boxes with no corresponding predictions are considered false negatives (FN). Precision and recall are defined as:

$$Precision = \frac{TP}{TP + FP} \tag{7}$$

$$Recall = \frac{TP}{TP + FN} \tag{8}$$

The Average Precision (AP) at a specific IoU threshold $\alpha$ is computed as the average of precision values across multiple recall levels:

$$AP_\alpha = \frac{1}{11} \sum_{r \in \{0.0, 0.1, \ldots, 1.0\}} Precision_\alpha(r) \tag{9}$$

Finally, mAP is the average of AP across multiple IoU thresholds:

$$mAP = \frac{1}{5} \sum_{\alpha \in \{0.3, 0.4, 0.5, 0.6, 0.7\}} AP_\alpha \tag{10}$$

This metric provides a comprehensive measure of both classification accuracy and localization performance of the object detection model.

*4.2. Contextual Hazard Detection Evaluation*

Following the evaluation of the object detection module, the contextual hazard reasoning capability of the proposed framework was assessed using a dedicated dataset.

4.2.1. Contextual Hazard Evaluation Dataset

As presented in Table 3, this study focuses on four contextual hazard categories relevant to construction site safety. Subsequently, corresponding construction site images representing these hazard conditions were collected to construct the contextual hazard detection dataset. A portion of these images originates from historical inspection records captured by site safety managers. Due to the limited availability of historical hazard images, the dataset was further expanded using publicly available construction safety dataset [57]. A systematic filtering process was applied to ensure dataset quality and relevance. This process included verifying that images corresponded to realistic construction site scenarios, removing low-quality or blurry images.

To generate initial hazard annotations and explanatory rationales, the GPT-4o VLM was employed. Each image was provided to the model together with structured prompts specifying the hazard definitions and output format used in this study. The model was configured with a temperature of 0.1 to encourage deterministic responses and a maximum output length of 180 tokens to ensure concise explanations. The model produced hazard labels selected from the predefined categories along with short textual rationales describing the identified safety conditions.

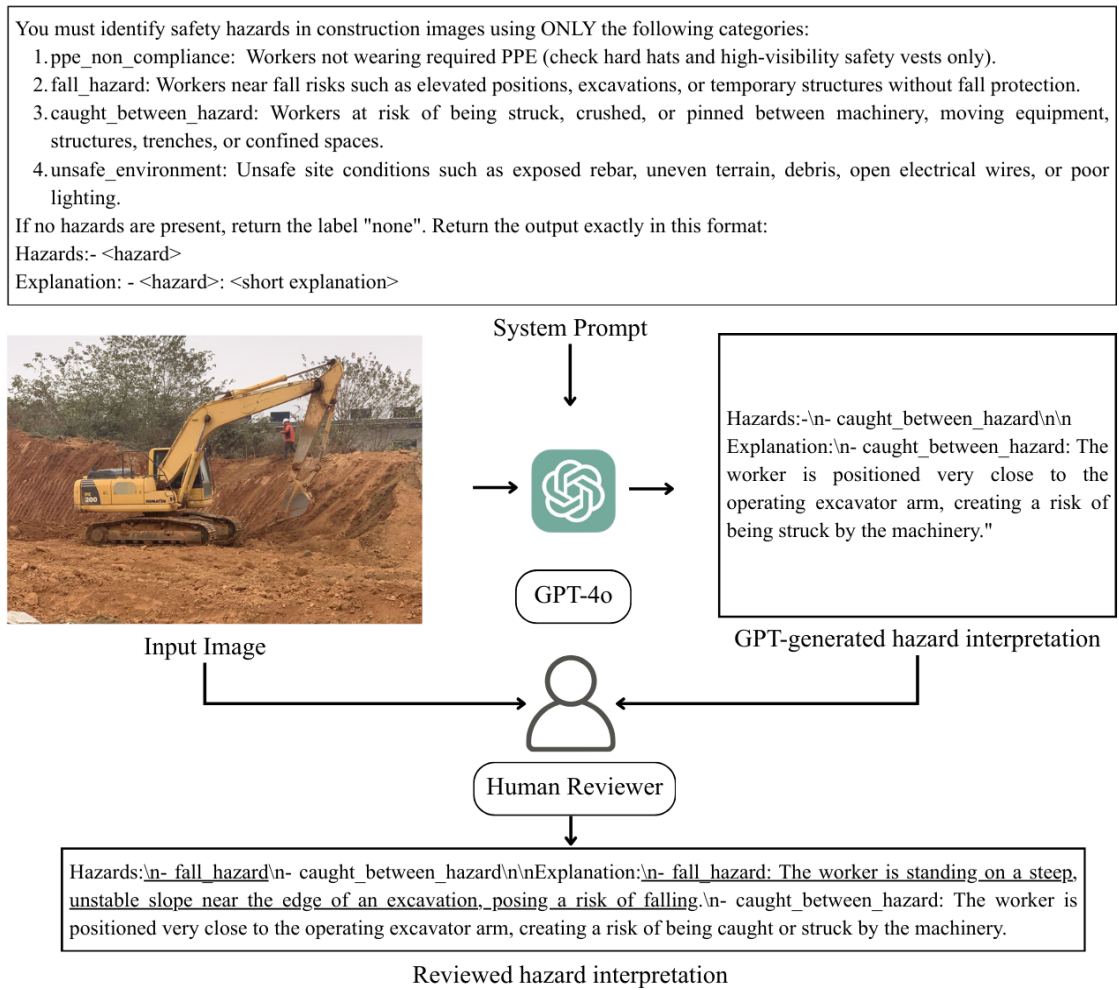

Figure 3 – Hazard annotation generation and validation workflow

Following automatic generation, a trained annotator conducted a validation and refinement process to ensure the reliability of the annotations. A standardized evaluation rubric was applied to reduce subjective bias and maintain annotation consistency. The validation focused on two criteria: (1) the correct identification of hazards according to established occupational safety principles, and (2) the clarity and specificity of the accompanying explanations. Annotations containing incorrect hazard classifications, vague descriptions, or inconsistent reasoning were revised to ensure alignment with construction safety guidelines and dataset quality standards. After filtering, generation, and validation, the final dataset comprised 1,833 construction site images depicting various contextual hazard scenarios.

### 4.2.2. Experimental Configurations

To evaluate the contextual hazard reasoning capability of the proposed framework, several sVLMs were tested within the evaluation pipeline. The evaluated models included Gemma-3 (4B), Qwen-3 (2B and 4B), InternVL-3 (1B, 2B, and 4B), and SmolVLM-2B. These models were selected because of their relatively small parameter sizes (≤4B parameters), which make them suitable for deployment in resource-constrained environments while still providing strong multimodal reasoning capabilities. All models were implemented using the Hugging Face Transformers library following the inference guidelines provided in their respective repositories.

Following the preparation of the contextual hazard evaluation dataset described in Section 4.2.1, experiments were conducted to evaluate the effectiveness of the proposed detection-guided prompting strategy. Two experimental configurations were designed to assess the impact of incorporating spatial detection information into the hazard detection process.

- *Baseline configuration:* In the baseline setting, pre-trained sVLM was applied directly in a zero-shot setting without integrating the object detection module. The models received the raw construction site images together with structured prompts describing the hazard evaluation task. The prompts instructed the models to analyze the scene and identify hazards from the predefined categories listed in Table 3 (PPE non-compliance, fall hazard, caught-between hazard, and unsafe environment). In this configuration, the models relied solely on visual information from the image without any explicit spatial guidance or entity descriptions.

- *Proposed Detection-Guided Configuration:* The trained YOLOv11n detector was first applied to each test image to generate bounding boxes and class labels for workers and machinery. This detection output was then encoded into structured natural language prompts that explicitly specified: (1) the locations of detected workers (e.g., "Worker w1: center=0.558,0.518, w2: center=0.590,0.514"), and (2) the types and locations of nearby machinery. These detection-guided prompts were passed to the sVLM along with the complete input images, directing model attention toward identified entities while preserving full visual context.

### 4.2.3. Evaluation Metrics

The contextual hazard detection capability of the evaluated VLMs was assessed using metrics that capture both (1) classification accuracy and (2) quality of the generated explanations. Specifically, precision, recall, and F1-score were used to evaluate hazard classification performance, while BERTScore F1 was used to measure the semantic quality and accuracy of the generated rationales.

For hazard classification, predictions were compared against the ground-truth hazard labels in the contextual hazard evaluation dataset. Since multiple hazard categories may be present in a single image, the evaluation was conducted at the image level, where a prediction was considered correct if the predicted hazard categories matched the ground-truth annotations. F1-score is calculated as:

$$F1 = \frac{2.Precision.Recall}{Precision + Recall} \tag{11}$$

The second metric was applied to evaluate the natural-language rationales generated by the sVLMs to justify hazard predictions. BERTScore is a semantic similarity metric that evaluates the alignment between generated and reference texts by comparing contextualized token embeddings. BERTScore operates at the representation level, utilizing pre-trained transformer-based models. specifically, BERT (Bidirectional Encoder Representations from Transformers), to compute similarity scores. It measures precision, recall, and F1 by calculating the cosine similarity between each token in the candidate and reference sequences, thereby capturing fine-grained semantic relationships, including synonymy and contextual equivalence.

The BERTScore precision, recall, and F1 are calculated as:

$$Precision = \frac{\sum_{t \in T_p} max_{g \in T_p} sim(t, g)}{|T_p|} \tag{12}$$

$$Recall = \frac{\sum_{g \in T_g} max_{g \in T_p} sim(g, t)}{|T_g|} \tag{13}$$

$$BERTScoreF1 = \frac{2.Precision.Recall}{Precision + Recall} \tag{14}$$

where $T_p$ and $T_g$ represent token sets in the predicted and ground truth rationales, respectively, and *sim(t,g)* denotes cosine similarity between contextualized embeddings. The pre-trained 'roberta-large' model was used to generate token embeddings for BERTScore computation.

## 5. Results

### 5.1. Object Detection Results

The trained YOLOv11n model was evaluated on the testing split to establish its reliability in detecting workers and machinery. Figure 4 presents example detection outputs generated by the trained YOLOv11n model. Table 4 summarizes the detection performance across all object classes.

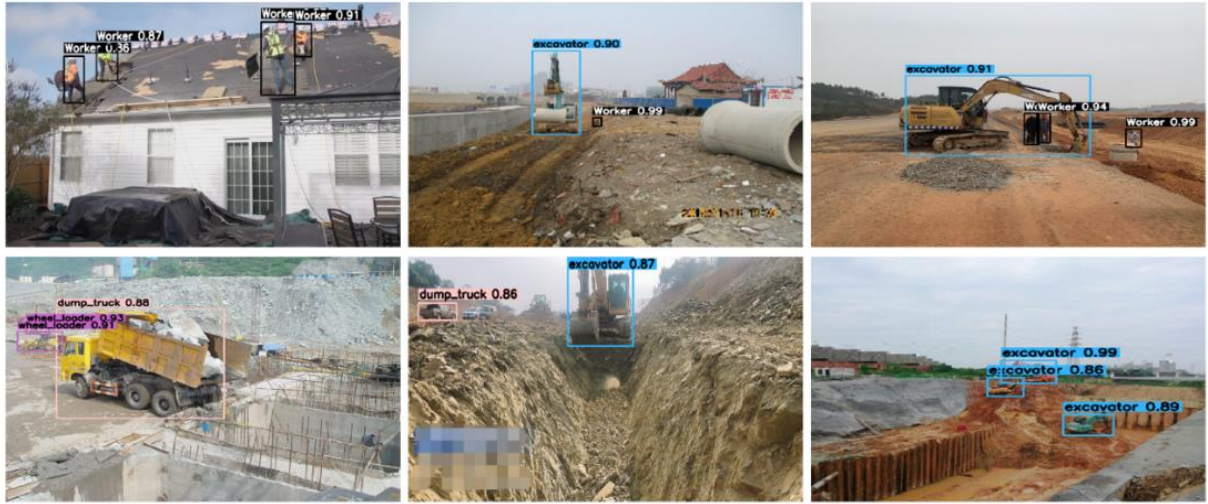

Figure 4 - Example object detection outputs

Table 4 - Object detection performance on the evaluation dataset

| Class | Precision | Recall | mAP@0.5 | mAP@0.5–0.95 |
|---|---|---|---|---|
| Worker | 85.6 | 87.5 | 90.5 | 74.0 |
| Backhoe loader | 94.7 | 96.4 | 98.9 | 88.9 |
| Cement truck | 92.9 | 94.6 | 97.4 | 84.1 |
| Compactor | 93.9 | 91.5 | 96.3 | 83.7 |
| Dozer | 93.1 | 91.6 | 95.1 | 80.1 |
| Dump truck | 74.8 | 73.6 | 79.6 | 56.8 |
| Excavator | 89.5 | 89.6 | 93.5 | 77.8 |
| Grader | 98.3 | 98.1 | 99.4 | 92.0 |
| Mobile crane | 85.3 | 86.4 | 92.3 | 71.8 |
| Tower crane | 83.8 | 76.7 | 80.5 | 60.9 |
| Wheel loader | 87.5 | 86.1 | 91.8 | 80.8 |
| **Overall** | **89.0** | **88.4** | **93.3** | **77.4** |

The YOLOv11n model achieved an overall mAP@0.5 of 93.3% and mAP@0.5:0.95 of 77.4%, indicating strong localization and classification performance across diverse construction object categories. Precision and recall were consistently high for most machinery classes (above 90%), showing the detector's ability to minimize both false positives and false negatives. Particularly strong results were observed for graders (mAP@0.5 = 99.4%) and backhoe loaders (mAP@0.5 = 98.9%), reflecting the model's robustness in detecting well-defined machinery. These results indicate that the detector provides sufficiently accurate bounding boxes and class labels to support the subsequent detection-guided reasoning module.

*5.2. Hazard Detection Results*

Figure 5 presents example outputs from the proposed hazard detection method using Gemma-3 4B. Hazard detection was evaluated by comparing the baseline and proposed configurations against the dataset.

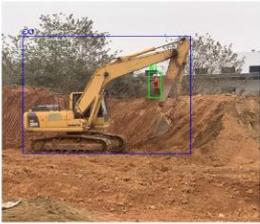
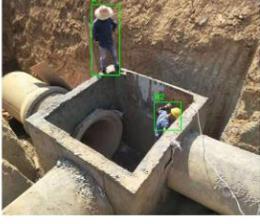
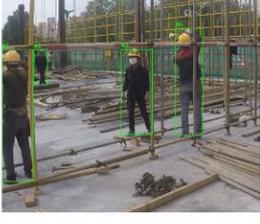
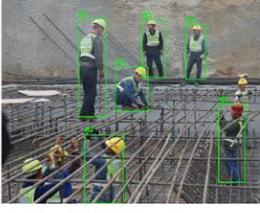

Figure 5 - Examples of hazard assessments generated by the proposed framework using YOLOv11n + Gemma-3 4B

5.2.1 Overall Hazard Detection Performance

Table 5 presents the overall hazard detection performance across the evaluated sVLMs for both configurations. Across all evaluated models, the proposed method consistently achieved higher F1-scores, demonstrating the effectiveness of incorporating spatial grounding into the

vision–language reasoning process. The highest performance was achieved by Gemma-3 4B, with an F1-score of 50.6%, compared to 34.5% under the baseline configuration. Similarly, Qwen-3-VL 4B improved from 37.4% to 46.3%, while SmolVLM 2B increased from 26.2% to 40.6%. Even smaller models such as InternVL-3 1B showed substantial improvements, with the F1-score increasing from 19.4% to 34.2%. These improvements indicate that the proposed detection-guided framework significantly enhances hazard detection performance across different model architectures and parameter sizes.

Table 5 - Overall F1-scores for baseline sVLM and proposed detection-guided sVLM approach

| Model | Baseline (%) | Proposed (%) | Improvement (%) |
| --- | --- | --- | --- |
| Gemma-3 4B | 34.5 | 50.6 | **+16.1** |
| Qwen-3-VL 4B | 37.4 | 46.3 | **+8.9** |
| Qwen-3-VL 2B | 30.3 | 38.7 | **+8.4** |
| InternVL-3 2B | 28.7 | 41.6 | **+12.9** |
| InternVL-3 1B | 19.4 | 34.2 | **+14.8** |
| SmolVLM 2B | 26.2 | 40.6 | **+14.4** |

The detailed precision, recall, and F1-score comparisons are illustrated in Figure 6. The results show a substantial increase in precision for the proposed method. For instance, the precision of Gemma-3 4B increased from 24.5% to 60.1%, while Qwen-3-VL 4B improved from 27.1% to 58.4%. These improvements indicate that the proposed approach significantly reduces false-positive hazard predictions.

In contrast to precision, recall values were generally higher in the baseline configuration. For example, Gemma-3 4B achieved a recall of 57.0% under the baseline setting but 43.7% under the proposed framework. This decrease in recall can be attributed to the more constrained reasoning process introduced by detection-guided prompting, which encourages the models to focus on spatially grounded hazards rather than producing broader hazard predictions.

Overall, the proposed detection-guided approach improved hazard classification performance across all models, with absolute F1-score increases between 8% and 16%. These findings confirm that grounding VLM reasoning in object-level detections reduces hallucinations, enhances contextual interpretation, and delivers more reliable results for near real-time hazard monitoring in construction environments.

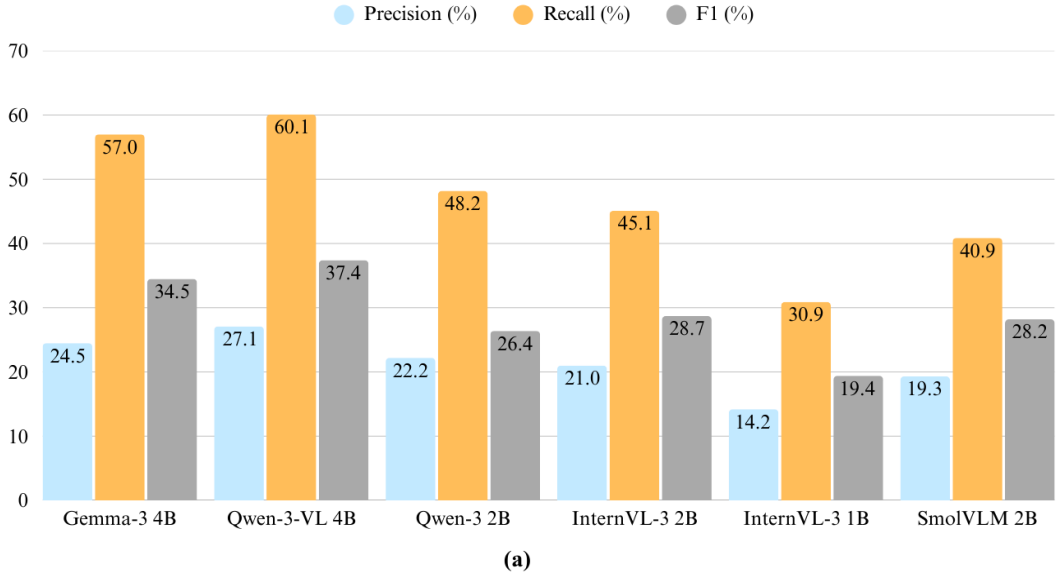

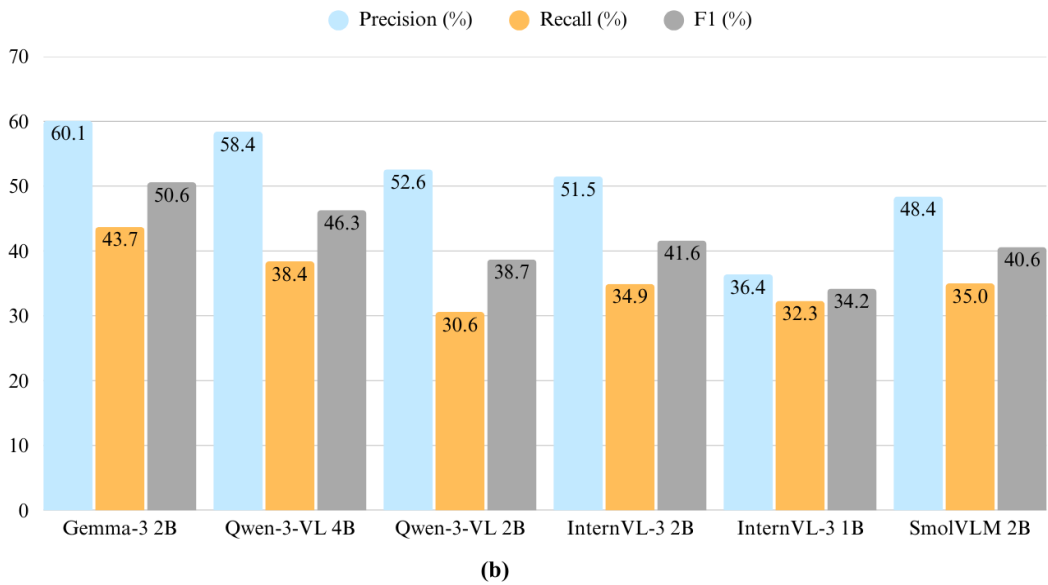

Figure 6 - Comparison of hazard classification performance for baseline sVLMs and proposed detection-guided sVLMs: (a) baseline configuration, (b) proposed method

## 5.2. Rationale Quality Results

The quality of rationales generated by the models was evaluated using BERTScore F1. Table 6 shows the results for each model under both baseline and proposed configurations.

Table 6 - Rationale quality results

| Model | Baseline | Proposed | Improvement |
| --- | --- | --- | --- |
| Gemma-3 4B | 0.63 | 0.82 | **+0.19** |
| Qwen-3-VL 4B | 0.56 | 0.80 | **+0.24** |
| Qwen-3-VL 2B | 0.57 | 0.78 | **+0.21** |
| InternVL-3 2B | 0.52 | 0.74 | **+0.22** |
| InternVL-3 1B | 0.46 | 0.62 | **+0.16** |
| SmolVLM 2B | 0.61 | 0.76 | **+0.15** |

The results demonstrate that the proposed detection-guided framework consistently improves the semantic quality of the generated hazard explanations across all evaluated models. The largest improvement is observed for Gemma-3 4B, where the BERTScore F1 increases from 0.6 to 0.82, indicating a substantial improvement in the alignment between generated and reference rationales. Similar improvements are observed for other models, including Qwen-3-VL 4B, which improves from 0.56 to 0.80, and InternVL-3 2B, which improves from 0.52 to 0.74. Overall, the rationale evaluation results demonstrate that the proposed framework not only improves hazard detection performance but also significantly enhances the interpretability and semantic accuracy of the generated explanations.

*5.4. Inference Efficiency*

In addition to evaluating hazard detection accuracy and rationale quality, the inference efficiency of the proposed framework was analyzed. Efficient inference is particularly important for practical deployment of construction safety hazard detection systems, where timely hazard identification is required for near-real-time decision support. Table 7 reports the average frames per second (FPS) achieved during the evaluation.

Table 7 - Inference efficiency of baseline vs detection-guided sVLMs

| **Model** | **Baseline (FPS)** | **Proposed (FPS)** |
|---|---|---|
| Gemma-3 4B | 0.43 | 0.42 |
| Qwen-3-VL 4B | 0.45 | 0.44 |
| Qwen-3-VL 2B | 0.42 | 0.43 |
| InternVL-3 2B | 0.42 | 0.40 |
| InternVL-3 1B | 0.48 | 0.48 |
| SmolVLM 2B | 0.45 | 0.45 |

The results indicate that the proposed detection-guided framework introduces minimal computational overhead compared to the baseline configuration. Across all evaluated models, the difference in inference speed between the two configurations is negligible, typically within 0.01–0.02 FPS. This small difference demonstrates that incorporating spatial grounding through object detection does not significantly impact the overall processing speed of the system.

Overall, the results demonstrate that the proposed detection-guided framework maintains comparable inference efficiency to baseline sVLM inference, while substantially improving hazard detection accuracy and explanation quality. These findings suggest that the framework is suitable for deployment in practical construction sites for hazard detection, where both accuracy and computational efficiency are critical.

## 6. Discussion

*6.1. Improvements in Hazard Detection and Rationale Generation*

The detection-guided framework consistently outperformed baseline sVLMs across all tested models. Overall, F1-scores improved by approximately 13–16 percentage points, with notable gains observed for smaller models such as InternVL-3 1B and SmolVLM-2B. The improvement was primarily driven by a substantial increase in precision, indicating that the detection-guided prompts helped reduce false-positive hazard predictions by grounding the reasoning process in explicitly detected workers and machinery.

Beyond classification, the framework enhanced the quality of rationales. Baseline models often produced vague or generic explanations, while detection-guided reasoning yielded clearer, context-specific justifications. This improvement in interpretability is critical for practical deployment, as inspectors require not only hazard alerts but also reliable reasoning to guide interventions.

*6.2. Computational Efficiency and Deployment Feasibility*

A key design goal of the framework was practical deployability in construction environments, where near real-time performance and hardware constraints are critical considerations. The experimental results show that the proposed detection-guided framework maintains comparable computational efficiency to the baseline configuration. Although both approaches operate below 1 FPS, the integration of the object detection module introduced only minimal additional overhead, primarily due to the lightweight YOLOv11n detector used to pre-extract workers and machinery from the scene.

Across all evaluated models, the inference speed remained within a narrow range of 0.40–0.48 FPS for both the baseline and detection-guided configurations. The YOLOv11n detection step added approximately 2.5 milliseconds per image, which had a negligible impact on the overall inference pipeline. This result indicates that incorporating spatial grounding through object detection does not significantly reduce throughput, allowing the system to maintain a similar level of computational performance.

These findings demonstrate that detection-guided small VLMs offer a balanced trade-off between reasoning capability and computational efficiency. While the detection-guided approach improves hazard identification accuracy and explanation quality, it does so without substantially increasing processing cost. This balance is particularly important for construction hazard detection, where systems must operate under practical hardware limitations.

*6.3. Limitations and Future Work*

Despite the promising results, several limitations should be acknowledged. First, the contextual hazard dataset used in this study focuses on a limited set of four hazard categories. These hazard categories, while most common, do not cover the full range of safety risks encountered on construction sites. Expanding the framework to include hazards such as electrical risks, chemical exposure, and ergonomic issues would improve its generalizability.

Another limitation is that the models were evaluated in a zero-shot setting without domain-specific fine-tuning. Although detection guidance improved reasoning, small VLMs still exhibited omission errors in cluttered, multi-hazard scenes. Future work should explore fine-tuning on construction-specific datasets, lightweight adaptation techniques such as LoRA, or knowledge distillation from larger VLMs to further enhance performance while maintaining efficiency.

Finally, evaluation relied on automated similarity metrics. Incorporating human-in-the-loop assessments by safety experts would provide a more practical measure of the usefulness and clarity of generated rationales in real-world inspection contexts.

**7. Conclusion**

This study introduced a detection-guided sVLM framework for contextual hazard detection in construction site environments. The proposed approach addresses key limitations of existing construction hazard detection systems, where traditional object detection methods lack contextual reasoning capabilities and vision–language models often struggle with spatial grounding and generate ambiguous or hallucinated interpretations. By integrating YOLOv11n-based object detection with sVLM reasoning through structured prompts, the framework enables hazard identification that is both spatially grounded and contextually informed.

Experimental evaluation demonstrated that the proposed detection-guided strategy significantly improves hazard detection performance across multiple sVLM architectures. Compared to the baseline zero-shot configuration, the proposed framework improved hazard detection F1-scores by approximately 13–16 percentage points, primarily due to substantial gains in precision while maintaining competitive recall levels. These results indicate that incorporating spatial information about workers and machinery helps reduce false-positive hazard predictions and allows models to reason more effectively about safety-critical interactions within construction scenes.

In addition to improving hazard detection accuracy, the proposed framework also enhanced the quality and interpretability of the generated explanations. The BERTScore evaluation showed consistent improvements across all evaluated models, with scores increasing from approximately 0.52–0.63 in the baseline configuration to 0.76–0.82 under the proposed framework. These improvements demonstrate that detection-guided prompting enables the models to produce explanations that are more semantically aligned with ground-truth safety rationales and more clearly describe the spatial conditions leading to potential hazards.

Overall, the findings of this study demonstrate that combining object-level spatial grounding with multimodal reasoning provides an effective approach for improving automated construction safety hazard detection. The proposed framework is model-agnostic, improves both prediction accuracy and explanation quality, and maintains efficient inference performance, making it suitable for integration into practical construction site monitoring platforms.

Future research will focus on expanding the framework to cover a broader range of construction hazards, incorporating domain-specific fine-tuning of vision–language models, and extending the approach to temporal reasoning for video-based safety monitoring, enabling more robust hazard detection in dynamic construction environments.

**Declaration of Competing Interest**

The authors declare that they have no known competing financial interests or personal relationships that could have appeared to influence the work reported in this study.

**Acknowledgement**

This work was supported by the NSERC Alliance under Grant ALLRP 576826 – 22.

**Data availability**

Data will be made available on request.

**Declaration of AI-assisted Technologies in the Writing Process**